\documentclass[10pt,twocolumn,letterpaper]{article}

\usepackage{cvpr}              
\usepackage{subcaption}
\usepackage{tikz}
\usetikzlibrary{positioning,arrows.meta}
\definecolor{cvprblue}{rgb}{0.21,0.49,0.74}
\usepackage[pagebackref,breaklinks,colorlinks,allcolors=magenta]{hyperref}
\usepackage{makecell}
\def\paperID{*****} 
\def\confName{CVPR}
\def\confYear{2026}

\title{LoViF 2026 Challenge on Real-World All-in-One Image Restoration: Methods and Results}

\author{Xiang Chen\textsuperscript{$*$} \quad Hao Li\textsuperscript{$*$} \quad  Jiangxin Dong\textsuperscript{$*$} \quad Jinshan Pan\textsuperscript{$*$} \quad Xin Li\textsuperscript{$*$} \quad Xin He \quad Naiwei Chen \\ Shengyuan Li \quad Fengning Liu \quad Haoyi Lv \quad Haowei Peng \quad Yilian Zhong \quad Yuxiang Chen \\ Shibo Yin \quad Yushun Fang \quad Xilei Zhu \quad Yahui Wang \quad Chen Lu \quad Kaibin Chen \quad Xu Zhang \\ Xuhui Cao \quad Jiaqi Ma \quad Ziqi Wang \quad Shengkai Hu \quad Yuning Cui \quad Huan Zhang \quad Shi Chen \\ Bin Ren \quad Lefei Zhang \quad Guanglu Dong \quad Qiyao Zhao \quad Tianheng Zheng \quad Chunlei Li \\ Lichao Mou \quad Chao Ren \quad Wangzhi Xing \quad Xin Lu \quad Enxuan Gu \quad Jingxi Zhang \quad Diqi Chen \\ Qiaosi Yi \quad Bingcai Wei \quad Mingyu Liu \quad Pengyu Wang \quad Ce Liu \quad Miaoxin Guan \\ Boyu Chen \quad Hongyu Li \quad Jian Zhu \quad Xinrui Luo \quad Ziyang He \quad Jiayu Wang \quad Yichen Xiang \\ Huayi Qi \quad Haoyu Bian \quad  Yiran Li \quad Sunlichen Zhou 
}

\begin{document}
\maketitle
\renewcommand{\thefootnote}{}
\footnotetext{$^{*}$X. Chen(\textcolor{magenta}{chenxiang@njust.edu.cn}), H. Li(\textcolor{magenta}{haoli@njust.edu.cn}), J. Dong(\textcolor{magenta}{jxdong@njust.edu.cn}), and J. Pan(\textcolor{magenta}{jspan@njust.edu.cn}) are the challenge organizers.}

\footnotetext{$^{*}$X. Li(\textcolor{magenta}{xin.li@ustc.edu.cn}) is the workshop organizer in assisting the organization of this challenge.}

\footnotetext{The other authors are participants of the LoViF 2026 Challenge on Real-World All-in-One Image Restoration.}


\begin{abstract}
This paper presents a review for the LoViF Challenge on Real-World All-in-One Image Restoration.
The challenge aimed to advance research on real-world all-in-one image restoration under diverse real-world degradation conditions, including blur, low-light, haze, rain, and snow. 
It provided a unified benchmark to evaluate the robustness and generalization ability of restoration models across multiple degradation categories within a common framework. 
The competition attracted 124 registered participants and received 9 valid final submissions with corresponding fact sheets, significantly contributing to the progress of real-world all-in-one image restoration.
This report provides a detailed analysis of the submitted methods and corresponding results, emphasizing recent progress in unified real-world image restoration. 
The analysis highlights effective approaches and establishes a benchmark for future research in real-world low-level vision.
%
%
%
\end{abstract}
    
\section{Introduction}
\label{sec:intro}
All-in-one image restoration aims to develop a unified model that can handle diverse types of image degradations within a single framework, such as rain, haze, noise, blur, and low-light conditions. Unlike traditional task-specific methods that are carefully designed and optimized for a single degradation type, all-in-one approaches emphasize generalization and scalability by learning representations across multiple image restoration tasks~\cite{chen2025foundir,jiang2025survey}. 

Compared to most existing methods that construct all-in-one restoration models by simply combining multiple synthetic datasets for training~\cite{Airnet,PromptIR,DiffUIR}, real-world scenarios present fundamentally different challenges due to the inherent domain gap between synthetic degradations and real-world datasets.
Recently, Li et al.~\cite{li2024foundir} proposed FoundIR, contributing a million-scale dataset for training robust all-in-one image restoration foundation models. By carefully adjusting internal camera settings and external imaging conditions, they are able to capture well-aligned image pairs through a well-designed data acquisition system across multiple rounds, together with a dedicated data alignment criterion. As a result, this dataset exhibits two notable advantages over existing training data: larger-scale real-world samples and higher-diversity data types.

To advance the development of real-world all-in-one image restoration, we organized the LoViF 2026 Challenge on Real-World All-in-One Image Restoration, in collaboration with the LoViF 2026 Workshops. This challenge aims to establish a practical and comprehensive benchmark for evaluating and enhancing the capability of all-in-one image restoration in real-world scenarios. To facilitate this challenge, we manually curate a small-scale subset from the original million-scale FoundIR dataset to construct the benchmark dataset, termed FoundIR-LoVIF.

This challenge is held with the LoViF Workshop~\footnote{\url{https://lovif-cvpr2026-workshop.github.io/}}, containing series of challenges on: real-world all-in-one image restoration, efficient VLM for multimodal creative quality scoring~\cite{lovif2026MQualityScoring}, weather removal in videos~\cite{lovif2026WeatherRemoval}, holistic quality assessment for 4D world model~\cite{lovif2026HQA}, and human-oriented semantic image quality assessment~\cite{lovif2026SeIQA}.
\section{LoViF 2026 Real-World All-in-One Image Restoration Challenge}
\label{sec:challenge}
The challenge focuses on real-world all-in-one image restoration, aiming to advance low-level vision research under diverse degradation conditions. 
Unlike conventional methods~\cite{chen2023learning,kong2023efficient} that address a single degradation type, this challenge emphasizes a unified setting in which restoration models are expected to handle multiple degradation categories within a common framework in real-world scenes.

\noindent\textbf{Dataset.} To benchmark the restoration performance in realistic scenarios, the dataset is provided by FoundIR~\cite{li2024foundir} and WeatherBench~\cite{guan2025weatherbench}, and is specifically designed to cover five common categories of real-world image degradation, \ie, \textit{blur, low-light, haze, rain, and snow}. 
The provided dataset includes diverse scenes for training, validation, and testing. 
Specifically, each degradation contains 4,900 paired low-quality and ground-truth image pairs of resolution $512 \times 512$ for training, together with 100 validation images and 100 test images. 
In total, the training set consists of 24,500 paired degraded images and their corresponding ground-truth references, while the validation and test sets each contain 500 degraded images. 

\noindent\textbf{Evaluation protocol.} To better quantify the rankings, we followed the scoring function from~\cite{li2025ntire} for three evaluation metrics in this challenge: PSNR, SSIM, and LPIPS.
The final ranking will be determined by a composite score that balances pixel-wise accuracy and perceptual quality. The weighted formula for the initial results is:
\begin{equation}
    \text{Score}= \text{PSNR(Y)} + 10 \times \text{SSIM(Y)} -5\times \text{LPIPS},
\end{equation}
where $\text{(Y)}$ denotes the PSNR and SSIM are measured with the $\text{Y}$ channel after coverting the image from RGB space to the YCbCr space. 
For LPIPS, we first normalize the image pixel values into the range $[-1,1]$, then utilize the Alex network configuration for distance measurement between restored images and ground-truth images.

By establishing such a dataset and protocol, the challenge aims to promote the development of robust restoration methods for real-world degraded images.
\section{Challenge Results}
\label{challenge_results}
The results of the LoViF 2026 Real-World All-in-One Image Restoration Challenge demonstrate a competitive landscape with several strong approaches. The challenge results are presented in Table~\ref{tab:results_1}. We report the performances of teams that submitted their fact sheets. The top three teams are HJHK-ClearVision (led by Xin He), RedMediaTech (led by Yilian Zhong), and \%sIR (led by Kaibin Chen), achieving final scores of 33.86, 33.58, and 32.63, respectively. These leading methods show clear performance advantages, though the margin among the top competitors remains relatively small, indicating a highly competitive benchmark. Other teams, such as GKD\_IR and DGL-team, also achieve comparable performance, further highlighting the overall progress in this field.

Despite these encouraging results, the challenge of real-world all-in-one image restoration remains far from solved. 
The variation in model complexity (in terms of GFLOPs and parameters) and the noticeable performance gap across teams suggest that designing efficient and robust models for diverse real-world degradations is still difficult. There is substantial room for improvement, particularly in balancing restoration quality with computational efficiency, and in enhancing generalization across complex, real-world scenarios.

\begin{table*}[]
\caption{Quantitative results of the LoViF 2026 Real-World All-in-One Image Restoration Challenge.}
\resizebox{\textwidth}{!}{
\begin{tabular}{c|c|c|c|cc}
\hline
~~~~Rank~~~~ & ~~~~Team Name~~~~        & ~~~~Team Leader~~~~  & ~~~~Final Score~~~~ & ~~~~GFLOPs (G)~~~~ & ~~~~Params (M)~~~~ \\ \hline
1    & HJHK-ClearVision & Xin He       & 33.86       & —          & —          \\
2    & RedMediaTech     & Yilian Zhong & 33.58       & —          & —          \\
3    & \%sIR            & Kaibin Chen  & 32.63       & 161.32     & 104.42     \\
4    & GKD\_IR          & Xu Zhang     & 32.30       & 1004.48    & 15.86      \\
5    & DGL-team         & Guanglu Dong & 32.19       & —          & 42.6       \\
6    & GU-day Mate      & Wangzhi Xing & 31.42       & 68.074     & 1.606      \\
7    & AIOVision        & Mingyu Liu   & 31.05       & 149.18     & 16.5       \\
8    & ColdWind         & Ce Liu       & 23.29       & —          & —          \\
—    & LR               & Jian Zhu     & 32.62       & 252.32     & 78.33      \\ \hline
\end{tabular}
}
\label{tab:results_1}
\end{table*}

\section{Teams and Methods}
\label{sec:teams_and_methods_1}

\subsection{HJHK-ClearVision}

This team builds their solution based on the WaveMamba~\cite{zou2024wave} architecture for image restoration. Their model follows an encoder–decoder structure in which each stage is composed of two key modules: LFSSBlocks and HFEBlocks. Specifically, the numbers of LFSSBlocks across the three layers of the encoder and decoder are set to $\left [ 1,2,4\right ]$, while the number of HFEBlocks is consistently configured as $\left [ 1,1,1\right ]$. To balance computational efficiency and model representation capability, they set the number of attention heads to 8 and use a base channel width $C=32$.
They also incorporate additional training data to improve the generalization ability of their model. In addition to the dataset provided by the competition organizers, they utilize two external datasets, FoundIR~\cite{li2024foundir} and WeatherBench~\cite{guan2025weatherbench}, to enrich the diversity of degradation patterns and expand the training data distribution.

\noindent \textbf{Training details.}
They conduct all experiments on a single NVIDIA RTX 4090 GPU. Their model is optimized using the $L_1$ loss function with the AdamW optimizer. The initial learning rate is set to $5\times10^{-4}$ and gradually decays to $1\times10^{-7}$ following a cosine annealing schedule over 500k training iterations. To increase data diversity and improve robustness, they apply extensive data augmentation strategies, including random rotations of $90^{\circ}$, $180^{\circ}$, and $270^{\circ}$ random horizontal and vertical flips, and random cropping to generate $512\times512$ image patches.
To further improve performance, they adopt a pseudo-label based re-training strategy. Specifically, they first use the model obtained from the initial training stage to perform inference on the low-quality images from the test set. The resulting restored outputs are treated as pseudo-labels and combined with the original training data to form an expanded dataset. They then retrain the model using this augmented dataset, enabling the model to better adapt to the data distribution of the target test set and improving the final performance.

\noindent \textbf{Testing details.}
During inference, they directly feed the degraded input images into the trained WaveMamba-based restoration network to generate the restored results. The final model used for testing is obtained after the pseudo-label based re-training stage.

\subsection{RedMediaTech}
\begin{figure}[t]
  \centering
  \includegraphics[width=1.0\linewidth]{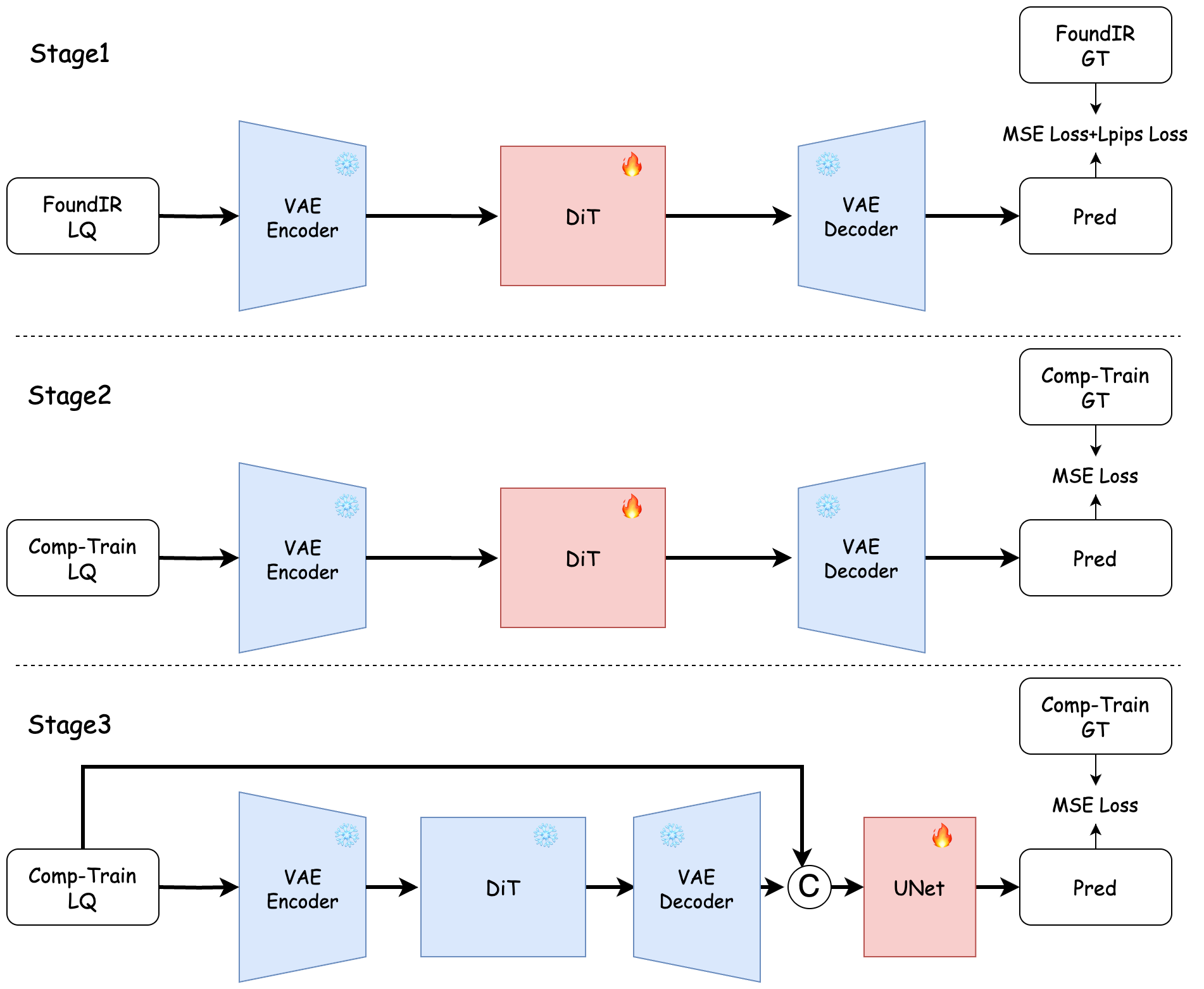}
  \caption{The overall framework of Team RedMediaTech.}
  \label{fig:net_team2_RedMediaTech}
\end{figure}

This team builds their solution on the pre-trained Stable Diffusion 3 Medium (SD3-Medium)~\cite{esser2024scaling} backbone and adopts a three-stage training strategy to progressively transfer the strong generative prior of SD3 to the target image restoration task. Their framework first learns general restoration capability from large-scale real-world data, then adapts to the competition domain through fine-tuning, and finally introduces a dedicated refinement module to enhance local structures and suppress remaining artifacts.
They initialize the restoration backbone from SD3-Medium and retain the pretrained VAE during training. Their design freezes the VAE parameters while optimizing only the DiT backbone, allowing the model to preserve the expressive latent representation learned during large-scale pretraining while adapting the denoising and restoration process to the target task. After the SD3-based backbone is fine-tuned on the competition dataset, they introduce an additional refiner module to further improve the restoration quality.
Their refiner is implemented using a UNet-based architecture similar to the U-Net used in Stable Diffusion 2.1. The refiner takes an input resolution of $512\times512$ with 6 input channels and 3 output channels in the RGB domain. Each block contains two convolutional layers, and the channel configuration is $(128,128,256,256,512,512)$. To improve feature aggregation, they incorporate one attention-based downsampling block and one attention-based upsampling block. During this stage, the SD3 backbone is completely frozen, and only the refiner parameters are optimized.
Their pipeline first processes a degraded input image through the SD3-based restoration backbone to produce an initial restored result. This intermediate restoration is then fed into the UNet refiner, which further enhances local textures and structural details. Their framework thus combines the strong generative prior of large-scale diffusion pretraining with the flexibility of a task-specific refinement module.
Their approach also utilizes additional training data. In particular, they employ FoundIR~\cite{li2024foundir}, a million-pair real-world image restoration dataset. This dataset is used only during the initial training stage to expose the model to diverse real-world degradations and improve the generalization capability of the SD3-based backbone.

\noindent \textbf{Training details.}
They train the model in three stages. In the first stage, they initialize the model with pretrained SD3-Medium weights and pretrain it on the FoundIR dataset. The VAE remains frozen while the DiT backbone is optimized. The learning rate is set to $1\times10^{-4}$, the batch size is 16, and the loss function combines MSE loss and LPIPS loss. This stage is trained for 80,000 iterations to learn general restoration priors.
In the second stage, they fine-tune the model on the competition training dataset to reduce the domain gap. The VAE is still frozen while the DiT backbone is fully optimized. The batch size is 8, the loss function is MSE loss, and the training runs for 110,000 iterations.
In the third stage, they append the UNet-based refiner after the SD3 backbone. During this stage, the entire SD3 backbone is frozen and only the refiner is trained on the competition dataset. The batch size is 8, the loss function is MSE loss, and the training lasts for 85,000 iterations.

\noindent \textbf{Testing details.}
They follow a two-step inference pipeline. The degraded input image is first restored using the SD3-based backbone to produce an initial restoration. This intermediate output is then passed through the refiner to obtain the final restored image. They do not utilize the degradation category labels available in the test set during inference.

\noindent \textbf{Ensembles and fusion strategies.}
They compute a Canny edge mask from the base image and use the refined result directly on edge regions to preserve high-frequency structures. On non-edge regions, they blend base and refined images by weighted averaging $I_{\text{out}}=(1-\alpha_{\text{nonedge}})I_{\text{refine}}+\alpha_{\text{nonedge}}I_{\text{base}}$ in RGB space, with optional edge dilation and feathering to ensure smooth transitions. This yields a 0.03 improvement in the score.

\subsection{\%sIR}
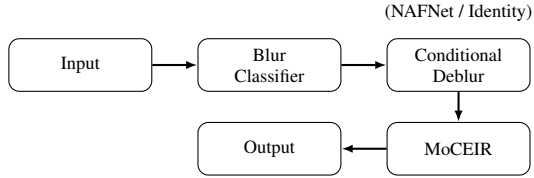
\begin{figure}[t]
\centering
\begin{tikzpicture}[
    node distance=4mm and 6mm,
    box/.style={
        draw,
        rounded corners,
        align=center,
        minimum height=7mm,
        minimum width=19mm,
        font=\scriptsize
    },
    arrow/.style={-{Latex[length=1.5mm]}, thick}
]
\node[box] (input) {Input};
\node[box, right=of input] (router) {Blur\\Classifier};
\node[box, right=of router] (deblur) {Conditional\\Deblur};
\node[box, below=of deblur] (restore) {MoCEIR};
\node[box, left=of restore] (output) {Output};

\draw[arrow] (input) -- (router);
\draw[arrow] (router) -- (deblur);
\draw[arrow] (deblur) -- (restore);
\draw[arrow] (restore) -- (output);
\node[font=\scriptsize, above=1mm of deblur] {(NAFNet / Identity)};
\end{tikzpicture}
\caption{The overall framework of Team \%sIR.}
\label{fig:net_team3_sIR}
\end{figure}

This team proposes a unified all-in-one image restoration framework built upon the MoCEIR~\cite{zamfir2025complexity} architecture, complemented by a conditional deblurring front-end. Their approach combines a frequency-guided mixture-of-experts restoration backbone with a lightweight blur-aware preprocessing module. The goal is to improve restoration performance on blur-heavy inputs while maintaining efficiency for other degradation types.
Their main restoration backbone is based on MoCEIR, which follows a U-shaped encoder–decoder architecture with frequency-guided mixture-of-experts routing. This design allows the model to dynamically allocate computation across experts according to the frequency characteristics of the input image. To better handle images dominated by motion or defocus blur, they introduce a conditional deblurring module before the main restoration network.
They design the deblurring front-end with two key components. The first component is a lightweight blur classifier built on ResNet18, which predicts a blur probability for each input image. This module acts as a routing gate that determines whether blur-specific preprocessing should be activated. The second component is a dedicated NAFNet~\cite{chen2022simple} deblurring branch. If the predicted blur probability exceeds a predefined threshold, the input image is first processed by the NAFNet branch before being sent to the MoCEIR backbone; otherwise, the original input image is directly forwarded to the restoration network. Their conditional design avoids unnecessary preprocessing for non-blurred images while improving restoration quality for blur-dominated cases.
Their overall restoration pipeline first normalizes the RGB input image to the $[0,1]$ range. The blur classifier then estimates the blur probability. When the probability exceeds a preset threshold, the image is processed by the NAFNet deblurring branch. The resulting image is subsequently fed into the MoCEIR backbone for unified restoration, and the final restored image is saved as the output.

\noindent \textbf{Training details.}
They train the entire framework in three stages. In the first stage, they fine-tune the pretrained MoCEIR backbone on the AIO dataset for 30 epochs using $256\times256$ patches and a batch size of 8 per GPU with bf16 mixed precision. They adopt a Muon/AdamW hybrid optimizer, where the Muon parameter group uses a learning rate of $1\times10^{-3}$ and the AdamW group uses $1\times10^{-4}$. The training includes three warm-up epochs and optimizes a combination of L1 loss, FFT loss, and a routing balance term.
In the second stage, they train the deblurring module on AIO data at a resolution of $512\times512$. The blur classifier is first trained as a binary classification model for 10 epochs with a batch size of 32, learning rate $1\times10^{-3}$, and weight decay $1\times10^{-4}$. Afterward, the NAFNet branch is trained on blur-specific data for 100 epochs using a batch size of 8, learning rate $1\times10^{-4}$, weight decay $1\times10^{-4}$, gradient accumulation of 2, and bf16 mixed precision.
In the final stage, they jointly fine-tune the deblurring module and the MoCEIR backbone as a unified system for 30 epochs using $256\times256$ patches and a batch size of 6 per GPU. The Muon learning rate for the backbone is $1\times10^{-3}$, the AdamW learning rate for the backbone parameters is $1\times10^{-4}$, and the deblurring branch is optimized with a smaller AdamW learning rate of $2\times10^{-5}$. Three warm-up epochs are also used in this stage.

\noindent \textbf{Testing details.}
They merge the blur classifier, the NAFNet deblurring branch, and the MoCEIR backbone into a single unified model for inference, resulting in one checkpoint of approximately 400 MB. During testing, the full $512\times512$ input image is processed directly without tiling. They do not employ test-time augmentation, self-ensemble, model ensemble, or other inference enhancements. Each image passes once through the pipeline of blur estimation, optional deblurring, unified restoration, and final output generation.

\subsection{GKD\_IR}
\begin{figure}[t]
  \centering
  \includegraphics[width=1.0\linewidth]{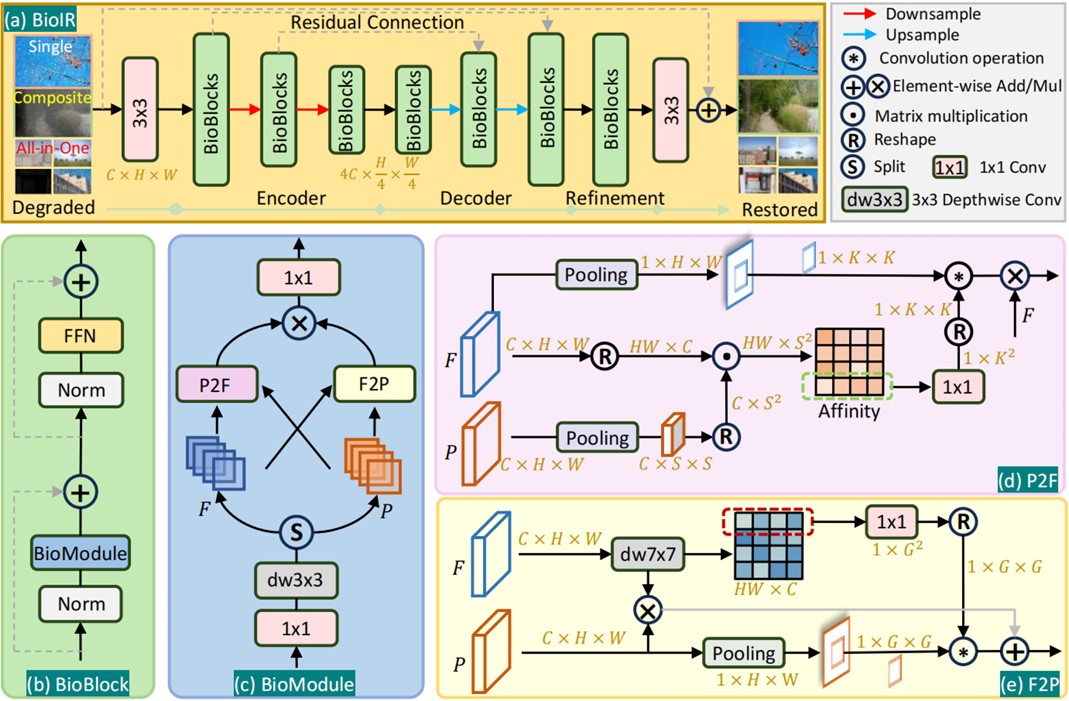}
  \caption{The overall framework of Team GKD\_IR.}
  \label{fig:net_team4_GKD_IR}
\end{figure}

This team proposes BioIR, an efficient and universal image restoration framework inspired by mechanisms of the human visual system. Their method introduces two bio-inspired modules~\cite{cuibio}, Peripheral-to-Foveal (P2F) and Foveal-to-Peripheral (F2P), to emulate the interaction between peripheral and foveal perception. By enabling dynamic interaction and recalibration between global contextual information and fine-grained spatial details, their framework is designed to handle single-degradation, all-in-one, and composite degradation scenarios while maintaining computational efficiency.
Their model adopts a plain U-shaped encoder–decoder architecture, where BioBlock serves as the fundamental building unit. Instead of using standard self-attention mechanisms, they design a BioModule that splits the feature representation into two complementary pathways. The P2F module propagates large-field contextual signals from peripheral regions to local spatial areas using pixel-to-region affinity, allowing the network to capture global context effectively. In parallel, the F2P module distributes fine-grained spatial details from focal regions to broader areas through a combination of element-wise modulation and dynamic convolution operations. Finally, the two feature streams are recalibrated through element-wise multiplication, enabling high-order feature interactions between contextual and local representations.

\noindent \textbf{Training details.}
They train the model using the Adam optimizer with an L1 loss computed in both the spatial and frequency domains. For the all-in-one restoration task, they adopt a patch size of $256\times256$, a batch size of 1, and an initial learning rate of $2\times10^{-4}$. Random horizontal and vertical flips are applied for data augmentation to improve generalization. The model is trained for a total of 200 epochs.

\noindent \textbf{Ensembles and fusion strategies.}
They do not employ explicit multi-model ensemble techniques. Instead, their framework incorporates an implicit fusion strategy within the BioModule design. Specifically, they fuse two complementary streams of visual information: large-field contextual signals from the peripheral pathway and fine-grained spatial details from the foveal pathway. These two representations are aggregated and recalibrated through an element-wise multiplication mechanism, allowing the network to effectively harmonize global context with local texture information.

\subsection{DGL-team}
\begin{figure}[t]
  \centering
  \includegraphics[width=1.0\linewidth]{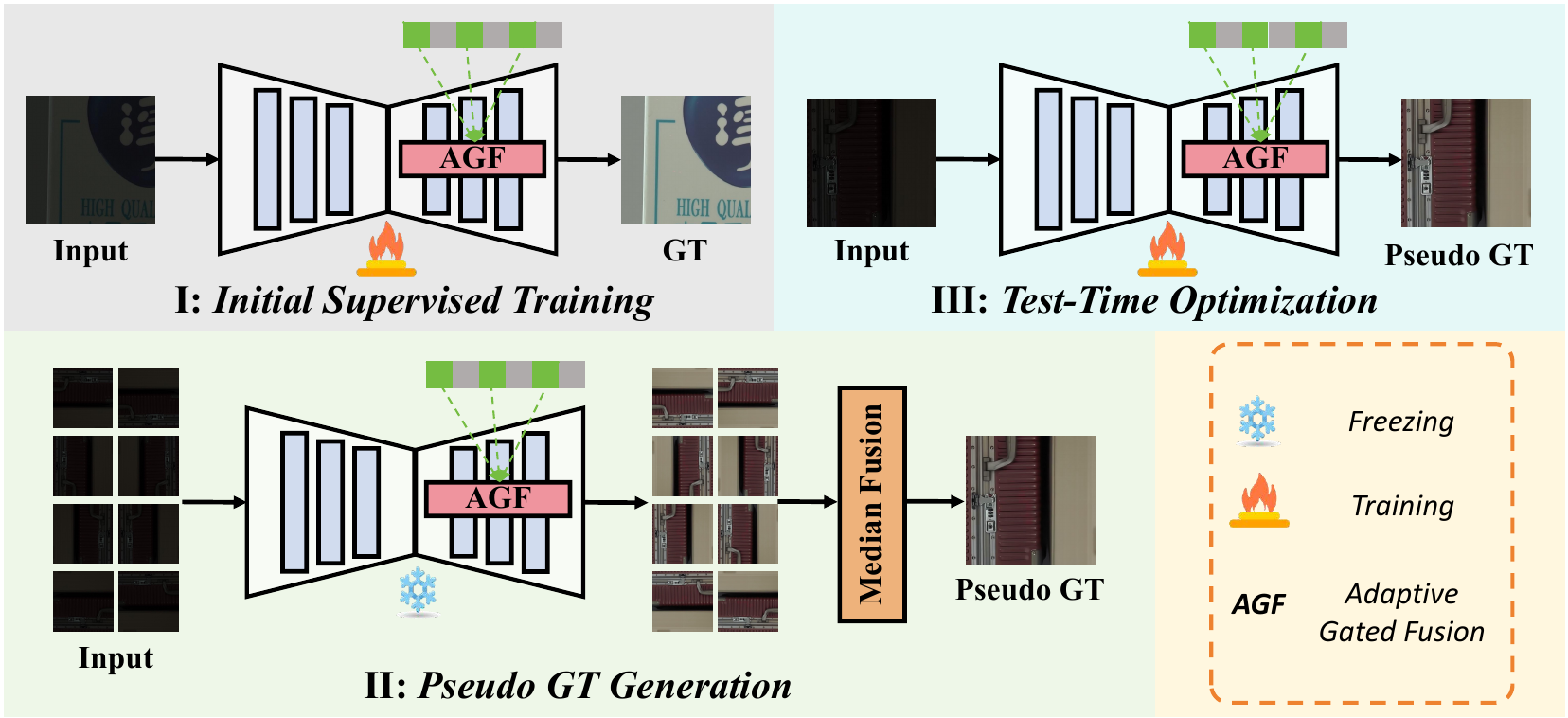}
  \caption{The overall framework of Team DGL-team.}
  \label{fig:net_team5_DGL-team}
\end{figure}

This team proposes a method built upon the recent all-in-one image restoration model DATPRL-IR~\cite{dong2026learning}. Since the current challenge focuses exclusively on natural image restoration, they simplify the original architecture by removing the domain prompt pool. The network mainly consists of an encoder-decoder-based image restoration backbone and a task prompt pool. The task prompt pool stores numbers of prompts, implicitly holding task-relevant prompt information. Given an input image, the network can adaptively select the most relevant prompts based on similarity and combine them at the instance level into a task prompt representation. The task prompt representation is then integrated into the decoder through gated cross-attention to guide the restoration process. This design allows instance-level prompt reuse and can effectively leverage both shared and specific task knowledge to guide restoration in multi-scenario images.
For the training phase, as shown in Fig.~\ref{fig:net_team5_DGL-team}, they adopt a two-stage training scheme, consisting of an initial supervised training stage and a subsequent pseudo-label-based test-time optimization stage.

\textit{Stage I: Initial supervised training.}
In the first stage, they train the simplified DATPRL-IR model using only the training set provided by the challenge organizers, without introducing any external data. They use only the L1 loss to obtain a strong initial restoration model. This stage aims to learn a reliable restoration mapping from the official training set and provide a stable initialization for the subsequent optimization stage.

\textit{Stage II: Pseudo-label-based test-time optimization.}
Starting from the model obtained in Stage I, inspired by Rong et al.~\cite{rong2025strrnet} and Li et al.~\cite{li2025ntire}, they further introduce a pseudo-label-based test-time optimization strategy to better adapt the model to the test data distribution. Specifically, they first use the Stage I model to generate restoration results for the test set. During inference, each test image is augmented into 8 geometrically transformed variants through horizontal flipping and rotation. The model then produces 8 corresponding restoration outputs. Different from existing self-ensemble methods~\cite{lim2017enhanced} that typically perform mean fusion, they adopt median fusion to aggregate these eight outputs into the final prediction used as the pseudo label. Compared with averaging, median fusion is more robust to outliers and unstable predictions, and can better suppress redundant or inconsistent responses among multiple augmented outputs.After obtaining the pseudo labels for the test set, they use them as supervision to fine-tune the Stage I model, resulting in their final model. In this stage, in addition to the L1 loss, they further introduce a perceptual loss~\cite{johnson2016perceptual} to improve perceptual quality.

\textit{Inference strategy.}
At test time, they use the model obtained after Stage II for final prediction. To maintain consistency between training and inference, they adopt a sliding-window inference strategy with a window size of $256\times256$, which matches the training crop size used in Stage II. To reduce boundary artifacts caused by patch-wise restoration, adjacent windows are processed with an overlap of 16 pixels. For the overlapping regions, they average the predicted pixel values to obtain a smooth transition between neighboring windows.

\noindent \textbf{Training details.}
Their method is implemented in PyTorch, and all experiments are conducted on a single NVIDIA GeForce RTX 5090 GPU. They use only the official dataset provided by the challenge, without any additional external data.
In Stage I, the model is trained using the $L_1$ loss only. They use the Adam optimizer with $\beta_1=0.9$ and $\beta_2=0.99$. The training patch size is set to $128\times128$, and the batch size is set to $6$. During data loading, they ensure that each mini-batch contains at least one sample from each degradation category, which helps alleviate imbalance across different degradation types. They apply standard data augmentation including random flipping and random rotation. The initial learning rate is set to $4\times10^{-4}$ and gradually decayed to $1\times10^{-6}$ using a cosine annealing schedule. The model is trained for 2,000,000 iterations in total.
In Stage II, the model is fine-tuned using the pseudo labels generated on the test set. They optimize the model with a combination of $L_1$ loss and perceptual loss, where the perceptual loss weight is set to $0.005$. The training patch size is increased to $256\times256$, the batch size is set to $1$, and the learning rate is fixed at $1\times10^{-5}$ throughout training. This stage is trained for 250,000 iterations.

\subsection{GU-day Mate}
\begin{figure}[t]
  \centering
  \includegraphics[width=1.0\linewidth]{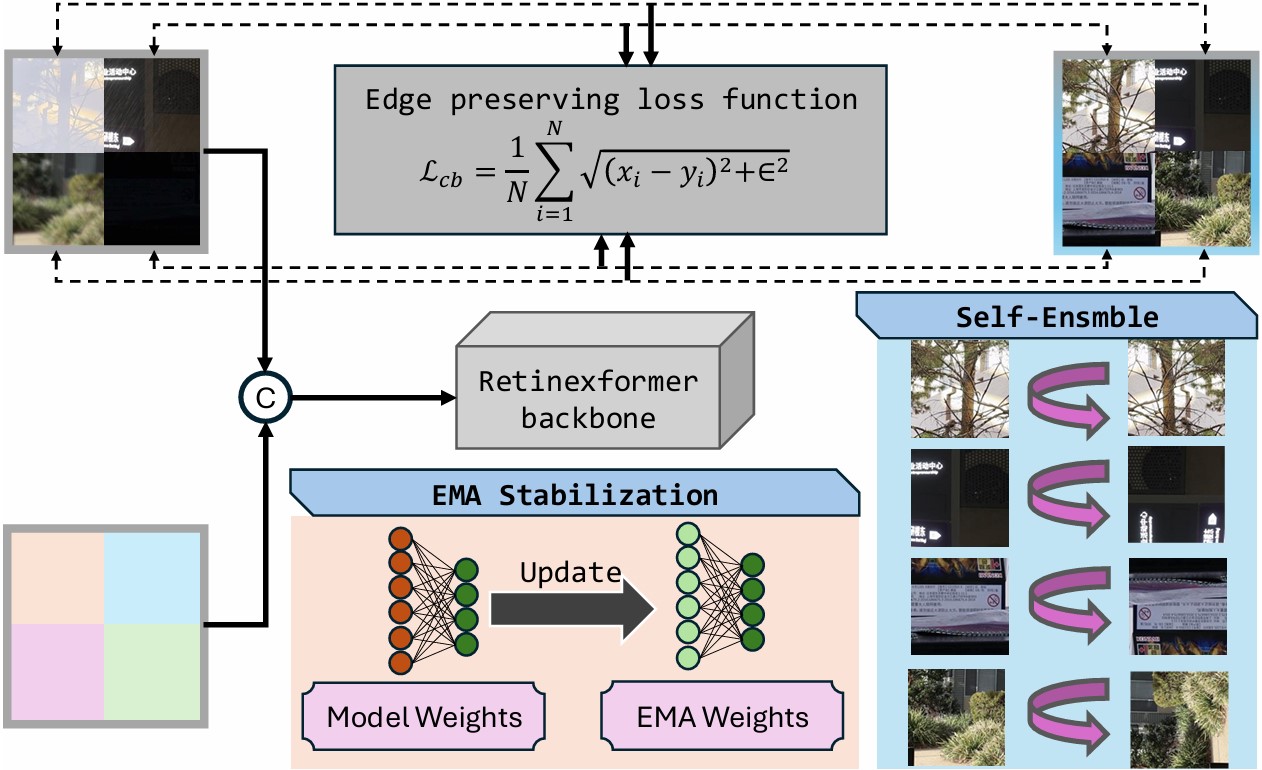}
  \caption{The overall framework of Team GU-day Mate.}
  \label{fig:net_team6_GUdayMate}
\end{figure}

This team builds their solution based on the Retinexformer architecture and introduces a curriculum-based multi-stage training strategy to progressively improve restoration capability. Their approach focuses on gradually expanding the receptive field during training, enabling the model to first learn low-level textures and then capture larger structural patterns at higher resolutions.
They adopt the Retinexformer network as the main restoration backbone and train it through a dynamic training pipeline that spans eight progressive stages over a total of 300 epochs. The training process starts with small image crops of $128\times128$ and a batch size of 8, allowing the model to focus on learning fine-grained local textures. As training progresses, the crop size is gradually increased while the batch size is reduced, eventually reaching a final resolution of $512\times512$ with a batch size of 1. This curriculum-style training scheme allows the model to progressively enlarge its effective receptive field and better handle global structural restoration at higher resolutions.
They also modify the optimization objective by replacing the standard L1 loss with Charbonnier loss ($\epsilon =10^{-3}$). This loss function provides edge-preserving gradients and remains differentiable near zero, which helps stabilize optimization and improves the preservation of fine details. To further improve training stability and generalization, they maintain an Exponential Moving Average (EMA) version of the model parameters with a decay factor of 0.99, producing a more stable set of weights for inference.

\noindent \textbf{Training details.}
They adopt a multi-stage curriculum training strategy that progressively increases the crop size from $128\times128$ to $512\times512$ over 300 epochs, while correspondingly reducing the batch size to balance memory usage and enlarge the effective receptive field for global restoration. To improve optimization stability and perceptual quality, they replace the standard L1 loss with Charbonnier loss, maintain an EMA version of the model with a decay of 0.99 during training, and apply self-ensemble TTA at inference for more robust predictions.

\noindent \textbf{Testing details.}
They execute the final inference for the GU-day Mate pipeline using the Retinexformer architecture, specifically loading the transductive fine-tuned weights to restore images from the LoViF final test set while applying self-ensemble augmentations to maximize visual fidelity.

\subsection{AIOVision}
\begin{figure}[t]
  \centering
  \includegraphics[width=1.0\linewidth]{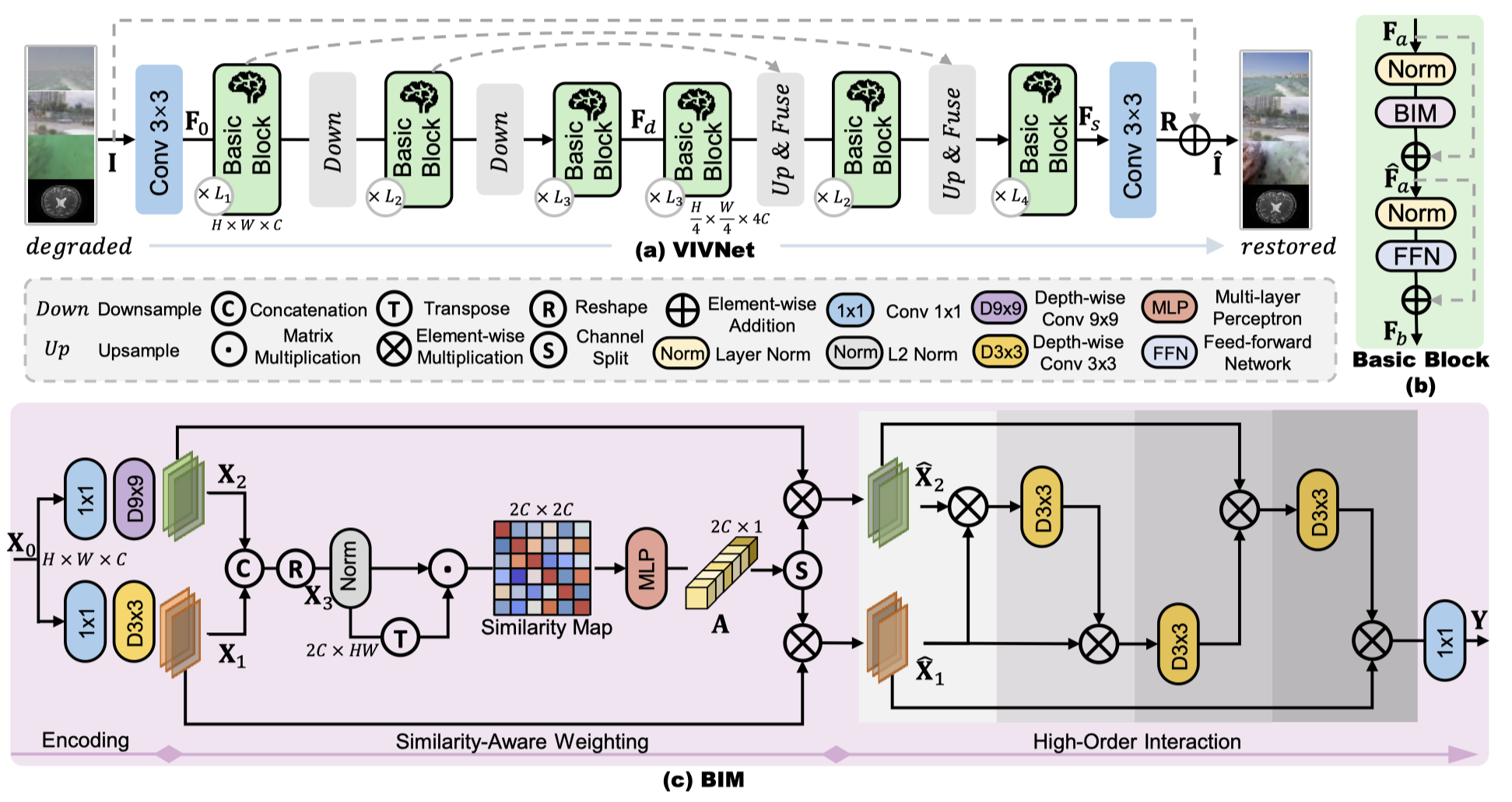}
  \caption{The overall framework of Team AIOVision.}
  \label{fig:net_team7_AIOVision}
\end{figure}

This team builds their solution based on VIVNet~\cite{cui2026visual}, which adopts a lightweight U-shaped architecture for image restoration. Their framework is designed to extract hierarchical visual features while maintaining computational efficiency. Within each block of the network, they introduce a brain-inspired module that models key perceptual mechanisms of the human visual system.
They design the encoding stage to capture multi-scale visual cues using depth-wise convolutions with different receptive fields, allowing the network to extract features at multiple spatial scales. To further enhance informative representations, they incorporate a similarity-aware weighting mechanism that adaptively emphasizes relevant features based on cosine similarity. In addition, they model high-order feature interactions through iterative element-wise multiplications combined with lightweight convolutional operations. This design enables the network to effectively integrate contextual and structural information during the restoration process.
Their main novelty lies in the integration of these brain-inspired components into a lightweight restoration framework. By combining multi-scale encoding, similarity-aware feature weighting, and high-order feature interactions within a U-shaped architecture, their method aims to achieve strong restoration performance while maintaining high computational efficiency across multiple restoration scenarios.

\noindent \textbf{Training details.}
They train the model using the AdamW optimizer with an initial learning rate of $2\times10^{-4}$ and a batch size of 16. The training process adopts a Linear Warmup + Cosine Annealing learning rate schedule, with 15 warm-up epochs followed by cosine decay. The model is trained for 150 epochs using image patches of size $128\times128$.

\noindent \textbf{Testing details.}
During training, they randomly select 100 images from the official training set to form a validation subset for performance evaluation and model selection. The final model obtained after training is used for inference on the test images.

\subsection{ColdWind}
This team proposes a solution that builds upon the UniRestore framework~\cite{chen2025unirestore}, a unified perceptual and task-oriented image restoration model using a diffusion prior. The original baseline extracts features using an autoencoder with a Complementary Feature Restoration Module (CFRM) and employs a Task Feature Adapter (TFA) to facilitate adaptive feature fusion in the decoder. To address the complex scenarios in real-world all-in-one image restoration efficiently and robustly, they introduced three critical improvements to the baseline.

\textit{Adaptive degradation control via prompt gating.}
They design and integrate a novel degradation intensity prediction head (deg-head) into the network, which is used to evaluate the severity of input degradation.
The predicted degradation intensity is then injected into the decoder through prompt gating, endowing the model with a "judge first, adaptively restore later" capability.
This enables the model to dynamically adjust the restoration strength according to the specific degradation level of each input image.

\textit{Identity regularization for clean inputs.}
To prevent the model from aggressively altering non-degraded or high-quality regions, they introduce an Identity Loss during training. They randomly replace the degraded input images with their corresponding High-Quality (HQ) ground truths with a probability of $p_{id} = 0.15$. The model is heavily penalized if it alters these clean inputs, guided by the Identity loss with a weight of $\lambda_{id} = 0.1$.

\textit{Accelerated diffusion prior (SD-Turbo).}
Instead of using the standard Stable Diffusion prior which requires multiple iterations, they replaced the pre-trained prior with SD-Turbo. This allows them to reduce the diffusion sampling steps to $1$, vastly accelerating the inference speed while maintaining high perceptual quality.

\noindent \textbf{Training details.}
The model is trained in a distributed manner across 2 GPUs. Regarding the training hyperparameters, the total batch size is set to 2, and the input image resolution (patch size) is uniformly cropped to $256 \times 256$ during training. In addition to the standard reconstruction loss, the optimization objective jointly incorporates the aforementioned Identity loss ($p_{id} = 0.1$) for supervision.

\subsection{LR}
This team proposes LoRA-IR, a unified image restoration framework designed to tackle five mixed degradations (blur, haze, low-light, rain, snow) without requiring prior knowledge of the degradation type. The network adopts NAFNet as its primary backbone. They leverage the CLIP model (openai/clip-vit-large-patch14-336) to perform real-time degradation classification on the input image. CLIP extracts a 384-dimensional degradation prior feature vector ($de\_prior$), which is injected into each encoder layer via an SE-like channel attention mechanism for conditional restoration. Crucially, for blur degradation, they dynamically activate a Kernel Estimation Network (KE-Net) based on Normalizing Flows and a Kernel Attention Module (KAM) to guide feature enhancement using the estimated pixel-wise blur kernels. This branch remains inactive for other degradations. Furthermore, standard convolutions throughout the network are replaced with Low-Rank Adaptation (LoRA) convolutions (Conv2dMix), enabling efficient parameter sharing and degradation-specific tuning.

\noindent \textbf{Training details.}
The model is implemented in PyTorch and trained on 8 RTX 4090 GPUs. They use the AdamW optimizer ($\beta_1 = 0.9, \beta_2 = 0.999$, weight decay $= 10^{-3}$). The initial learning rate is set to $5 \times 10^{-4}$ for the backbone and $10^{-3}$ ($2.0 \times$ multiplier) for the blur-specific modules. They utilize a Cosine Annealing learning rate scheduler ($T_{max} = 200,000, \eta_{min} = 10^{-7}$) with a 2,000-iteration warmup. The total batch size is 32 (4 per GPU), and the patch size is $256 \times 256$. Data augmentation includes random horizontal flips and rotations. They optimize the network using a PSNR Loss for 200,000 iterations. Exponential Moving Average (EMA) with a decay rate of 0.999 is applied to maintain model weights.

\section*{Acknowledgments}
This work has been partly supported by the National Natural Science Foundation of China (Nos. U22B2049, 62522606, 62272233, 62532004). We thank DARK HORSE Inc. and Nanjing ENTRO VISION Technology Co., Ltd., for sponsoring this challenge.
\appendix
\section{Teams and Affiliations}
\subsection*{Organizers}

\noindent  \textit{\textbf{Title:}} LoViF 2026 Challenge on Real-World All-in-One Image Restoration


\noindent  \textit{\textbf{Members:}}
Xiang Chen (\textcolor{magenta}{chenxiang@njust.edu.cn}), Hao Li, Jiangxin Dong, and Jinshan Pan


\noindent  \textit{\textbf{Affiliations:}}
Nanjing University of Science and Technology




\subsection*{HJHK-ClearVision}


\noindent  \textit{\textbf{Members:}}
\noindent   Xin He (\textcolor{magenta}{hexin6770@163.com}), Naiwei Chen, Shengyuan Li, Fengning Liu, Haoyi Lv, and Haowei Peng

\noindent  \textit{\textbf{Affiliations:}}
\noindent Naval Aviation University

\subsection*{RedMediaTech}

\noindent  \textit{\textbf{Members:}}
\noindent   Yilian Zhong (\textcolor{magenta}{zhongyilian@fudan.edu.cn}), Yuxiang Chen, Shibo Yin, Yushun Fang, Xilei Zhu, Yahui Wang, and ChenLu

\noindent  \textit{\textbf{Affiliations:}}
\noindent   Xiaohongshu Inc

\subsection*{\%sIR}


\noindent  \textit{\textbf{Members:}}
\noindent   Kaibin Chen (\textcolor{magenta}{qsz20241918@student.fjnu.edu.cn})

\noindent  \textit{\textbf{Affiliations:}}
\noindent   Fujian Normal University; Quanzhou Institute of Equipment Manufacturing, Chinese Academy of Sciences

\subsection*{GKD\_IR}


\noindent  \textit{\textbf{Members:}}
\noindent   Xu Zhang (\textcolor{magenta}{zhangx0802@whu.edu.cn}), Xuhui Cao, Jiaqi Ma, Ziqi Wang, Shengkai Hu, Yuning Cui, Huan Zhang, Shi Chen, Bin Ren, and Lefei Zhang

\noindent  \textit{\textbf{Affiliations:}}
\noindent   Wuhan University

\subsection*{DGL-team}


\noindent  \textit{\textbf{Members:}}
\noindent   Guanglu Dong (\textcolor{magenta}{dongguanglu@stu.scu.edu.cn}), Qiyao Zhao, Tianheng Zheng, Chunlei Li, Lichao Mou, and Chao Ren

\noindent  \textit{\textbf{Affiliations:}}
\noindent   Sichuan University

\subsection*{GU-day Mate}


\noindent  \textit{\textbf{Members:}}
\noindent   Wangzhi Xing (\textcolor{magenta}{w.xing@griffith.edu.au}), Xin Lu, Enxuan Gu, Jingxi Zhang, Diqi Chen, Qiaosi yi, and Bingcai Wei

\noindent  \textit{\textbf{Affiliations:}}
\noindent   Griffith University, Sensetime, Dalian University of Technology, Massey University, Hong Kong Polytechnic University, Wuhan University

\subsection*{AIOVision}


\noindent  \textit{\textbf{Members:}}
\noindent   Mingyu Liu (\textcolor{magenta}{mingyu.liu@tum.de}), Yuning Cui, and Pengyu Wang

\noindent  \textit{\textbf{Affiliations:}}
\noindent   Technical University of Munich

\subsection*{ColdWind}


\noindent  \textit{\textbf{Members:}}
\noindent   Ce Liu (\textcolor{magenta}{liuce382@foxmail.com}), Miaoxin Guan, Boyu Chen, and Hongyu Li

\noindent  \textit{\textbf{Affiliations:}}
\noindent   Guangdong University of Technology

\subsection*{LR}


\noindent  \textit{\textbf{Members:}}
\noindent   Jian Zhu (\textcolor{magenta}{saverm666@gmail.com}), Xinrui Luo, Ziyang He, Jiayu Wang, Yichen Xiang, Huayi Qi, Haoyu Bian, Yiran Li, and Sunlichen Zhou

\noindent  \textit{\textbf{Affiliations:}}
\noindent   University of Electronic Science and Technology of China
{
    \small
    \bibliographystyle{ieeenat_fullname}
    \bibliography{main}
}

\end{document}